\documentclass[11pt]{article}
\ifdefined\pdfinfoomitdate\pdfinfoomitdate=1\fi
\ifdefined\pdfsuppressptexinfo\pdfsuppressptexinfo=15\fi
\ifdefined\pdftrailerid\pdftrailerid{}\fi
\usepackage[preprint]{acl}
\hypersetup{pdfcreator={},pdfproducer={}}
\usepackage{times}
\usepackage{latexsym}
\usepackage[T1]{fontenc}
\usepackage[utf8]{inputenc}
\usepackage{microtype}
\usepackage{graphicx}
\usepackage{amsmath,amssymb}
\usepackage{booktabs,multirow,array,tabularx}
\usepackage{colortbl}
\usepackage{enumitem}
\usepackage{placeins}

\newcommand{\coauth}{$^\star$}
\newcommand{\correspond}{$^\dag$}

\title{Xetrieval: Explaining Dense Retrieval through Feature Elicitation}

\author{
Zhixin Cai$^{1}$\coauth{}, 
Jun Bai$^{2}$\coauth{}, 
Yang Liu$^{2}$\coauth{}, 
Jiaqi Li$^{2}$, 
Yichi Zhang$^{1}$,
Taichuan Li$^{1}$,\\ 
\textbf{Zhuofan Chen$^{1}$,} 
\textbf{Zixia Jia$^{2}$,} 
\textbf{Zilong Zheng$^{2}$\correspond{},} 
\textbf{Wenge Rong$^{1}$}
\\[0.5ex]
\textsuperscript{1}School of Computer Science and Engineering, Beihang University \\
\textsuperscript{2}State Key Laboratory of General Artificial Intelligence, BIGAI 
}

\begin{document}
\maketitle
\begingroup
\renewcommand{\thefootnote}{}
\footnotetext{\coauth{} Equal contribution. \correspond{} Corresponding author.}
\endgroup
\begin{abstract}
Dense retrieval enables effective semantic matching, yet the semantic factors underlying its relevance scores remain opaque.
Sparse autoencoders (SAEs) support feature-level explanations by decomposing dense embeddings into sparse semantic features.
However, the document semantics underlying a retriever's relevance scores may not be sufficiently explicit in the embeddings to be recovered as interpretable SAE features.
We introduce Xetrieval, a framework for explaining dense retrieval through feature elicitation.
Feature Elicitors make retrieval-relevant aspects of document meaning more explicit through auxiliary views, facilitating their identification as interpretable SAE features.
A shared sparse dictionary learns from these varied representations to express query--document relations through describable features.
We assess their influence on relevance scores through interventions on the original document embeddings.
Experiments show that Xetrieval yields more interpretable features than an SAE baseline and identifies feature sets with a stronger influence on the retriever's scores.
Task-level feature modulation further demonstrates more effective retrieval steering.
These findings highlight the value of feature elicitation for relevance attribution and retrieval steering.%
\footnote{The source code is available at \url{https://github.com/Hihiczx/Xetrieval}.}
\end{abstract}

\section{Introduction}
\label{sec:introduction}

\begin{figure}[t]
    \centering
    \includegraphics[width=\columnwidth]{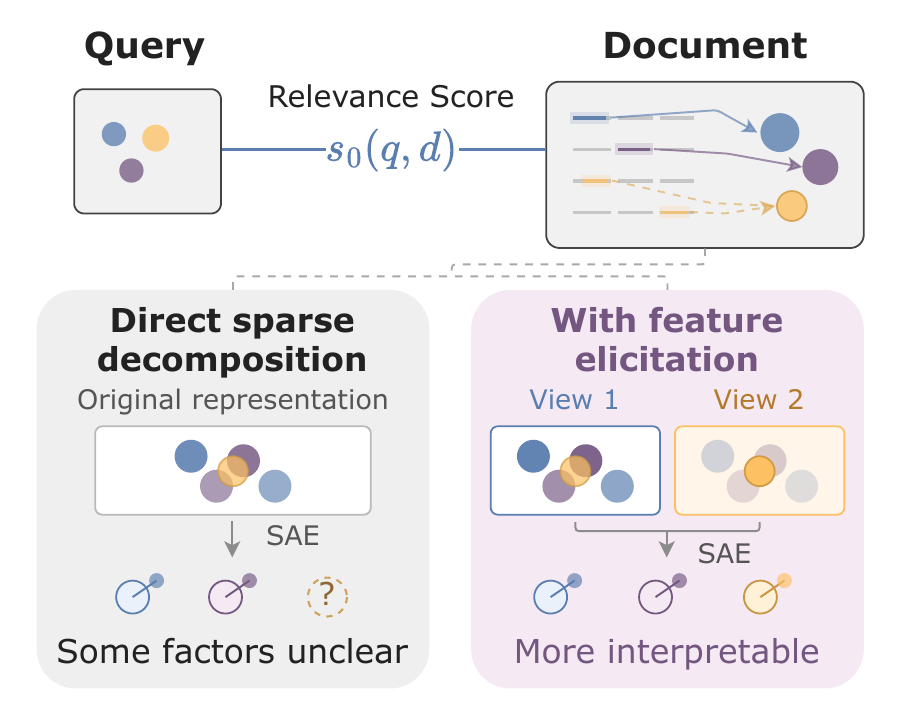}
    \caption{Illustration of feature elicitation for retrieval explanation. Elicited views (right) emphasize aspects of document meaning that can be difficult to recover as interpretable features through direct sparse decomposition (left).}
    \label{fig:motivation}
\end{figure}

Dense retrieval encodes queries and documents into a shared vector space and ranks documents by their similarity \citep{karpukhin2020dense}.
By modeling semantic relevance beyond term overlap, it helps address the vocabulary mismatch that limits lexical retrieval \citep{robertson2009probabilistic,berger2000bridging}.
This approach has demonstrated strong performance across diverse retrieval tasks and is widely used in modern retrieval systems \citep{karpukhin2020dense,shao2025reasonir}.

Despite these advances, the semantic basis of dense retrieval scores remains opaque \citep{opitz2025interpretable}.
Explaining dense retrieval involves identifying the semantic factors underlying a query--document score and expressing their role in terms people can understand.
Such explanations can support relevance attribution, targeted steering, and improvements to retrieval systems \citep{anand2022explainable}.
Token-level approaches provide lexical evidence for these scores, but their granularity makes it difficult to distinguish the semantic factors driving a retrieval decision and assess their respective contributions \citep{vasileiou-eberle-2024-explaining,ram-etal-2023-token}.

Sparse autoencoders (SAEs) provide a feature-level approach to retrieval explanation by decomposing dense embeddings into sparse features \citep{cunningham2023sparse/sae}.
These features can capture concepts across lexical variation, enabling relevance to be interpreted in terms of describable semantic factors rather than individual token contributions \citep{paulo2024automatically}.
By analyzing and modulating these feature activations, recent work has demonstrated effective explanation of query--document similarity and targeted steering of retrieval behavior \citep{kang2025interpret,park2025decoding}.

However, which semantic factors these methods recover as interpretable features depends on how they are expressed in the input representation.
Factors that influence model behavior can remain difficult to recover, but increasing their relative prominence can facilitate their recovery \citep{makelov2025towards}.
For dense retrieval, a document's central message, potential applications, and relevance to specific information needs can influence retrieval scores without being readily recovered as interpretable SAE features.
These meanings may be distributed across passages or left implicit in the text \citep{chen2025enrichindex,li2023parade}.
We therefore investigate auxiliary document views that consolidate the central content and make its applications and relation to information needs explicit, foregrounding these meanings for sparse feature learning and identification, as illustrated in Figure~\ref{fig:motivation}.

We introduce \textit{Xetrieval}, a framework for explaining dense retrieval through feature elicitation and sparse decomposition.
\emph{Feature Elicitors} express document semantics through auxiliary views that emphasize retrieval-relevant aspects of its content.
A shared sparse dictionary draws on these varied semantic expressions to provide a common vocabulary that connects queries and documents through shared features.
Natural-language descriptions of the shared features provide a feature-level account of the semantic connections between the query and document.

Our experiments show that Xetrieval yields more interpretable sparse features than an SAE baseline.
Pair-level interventions provide behavioral evidence for relevance attribution, showing that Xetrieval identifies feature sets with a stronger influence on the retriever's scores.
Task-level feature modulation further demonstrates more effective retrieval steering.
We also find that the elicited document views can improve retrieval performance.

\section{Related Work}
\label{sec:related_work}

\subsection{Explainable Dense Retrieval}
Token-level approaches explain similarity through interaction attributions \citep{moeller2023attribution,vasileiou-eberle-2024-explaining} or interpret retrieval embeddings through vocabulary-space projections \citep{ram-etal-2023-token}.
Structural probing examines the semantic information encoded in sentence-embedding subspaces \citep{nikolaev2023investigating}.
Related retrieval architectures make matching operations explicit through lexical weighting \citep{formal2021splade,lupart2026challenges}, token interactions \citep{khattab2020colbert}, or query--document modulation \citep{saxena-etal-2026-imrnns}.
Semantically structured embedding models associate representation components with explicit meaning aspects or natural-language questions \citep{opitz2022sbert,benara2024crafting}.

\subsection{Interpretable Sparse Features}
Motivated by superposition \citep{elhage2022toy}, SAEs decompose dense representations into sparse features \citep{cunningham2023sparse/sae} that admit natural-language descriptions.
For retrieval, \citet{park2025decoding} use SAE features to explain query--document similarity and support concept-based retrieval, while \citet{kang2025interpret} learn retrieval-oriented sparse features, interpret their semantics, and use their activations to steer retrieval.
Xetrieval focuses on the semantic presentation of documents, using auxiliary views to improve the learning and identification of interpretable sparse features.

\subsection{Semantic Reformulation for Retrieval}
Generated texts can make semantic content more explicit for relevance matching, supporting document expansion \citep{nogueira2019document,chen2025enrichindex}, synthetic retrieval supervision \citep{shao2025reasonir}, and query rewriting \citep{qin2025tongsearch}.
The embeddings of generated hypothetical documents can also represent queries for dense retrieval \citep{gao2023precise}.
Embedding-space approaches learn from textual supervision to approximate query reformulations \citep{zhang2025your} or align latent representations with explicit reasoning \citep{jin2026laser}.
In Xetrieval, textual reformulations guide the elicitation of interpretable features for retrieval explanation.

\section{Xetrieval: Feature Elicitation for Retrieval Explanation}
\label{sec:method}

\begin{figure*}[t]
    \centering
    \includegraphics[width=\textwidth]{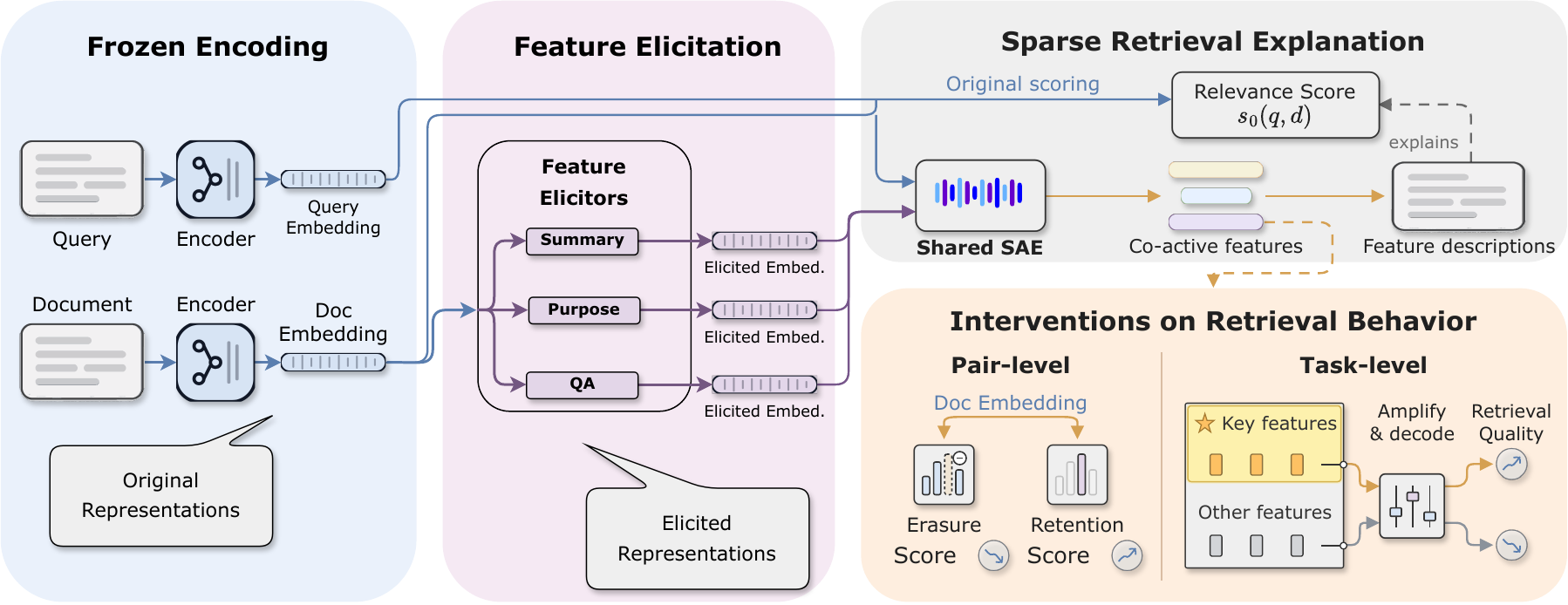}
    \caption{Overview of Xetrieval. Feature Elicitors make document semantics more explicit through auxiliary views, and a shared SAE identifies describable features shared by queries and documents. Pair-level interventions assess their influence on relevance scores. Task-level feature modulation evaluates retrieval steering with SAE-decoded representations.}
    \label{fig:overview}
\end{figure*}

Let $E_Q$ and $E_D$ be the frozen query and document encoders, producing $\mathbf q=E_Q(q)$ and $\mathbf z_d=E_D(d)$ in $\mathbb R^m$.
For a query--document pair, the explanation target is the retrieval score
\begin{equation}
    s_0(q,d)=\frac{\mathbf q^\top\mathbf z_d}
    {\|\mathbf q\|_2\,\|\mathbf z_d\|_2}.
    \label{eq:original_score}
\end{equation}
Xetrieval explains this score through natural-language descriptions of sparse features shared by the query and the document.
As illustrated in Figure~\ref{fig:overview}, Feature Elicitors make document semantics more explicit for sparse decomposition, while a shared sparse dictionary provides a common vocabulary for identifying this overlap across the document representations.

\subsection{Feature Elicitors}
\label{sec:projectors}

Following \citet{chen2025enrichindex}, we adopt three document perspectives $\mathcal T=\{\textsc{Summary},\textsc{Purpose},\textsc{QA}\}$ to guide feature elicitation.
\textsc{Summary} consolidates the document's main content into a concise account.
\textsc{Purpose} articulates the potential uses of that content.
\textsc{QA} reformulates document content as question--answer pairs, making its relevance to potential information needs explicit.
Feature Elicitors use these perspectives to make retrieval-relevant aspects of document meaning more explicit, facilitating their identification and interpretation through sparse decomposition.

For each training document $d$, an LLM generates a text $r_t(d)$ for each perspective $t$.
We encode each generated text as $\mathbf z_d^{(t)}=E_D(r_t(d))$.
To reduce the computational cost and latency of generating and encoding these texts during retrieval explanation, we train a \emph{Feature Elicitor} $P_t:\mathbb R^m\rightarrow\mathbb R^m$ to approximate $\mathbf z_d^{(t)}$ directly from $\mathbf z_d$.
Its output $\hat{\mathbf z}_d^{(t)}=P_t(\mathbf z_d)$ is the \emph{elicited view}.
We instantiate each elicitor as a one-hidden-layer network with $\tanh$ activation and $\ell_2$-normalized output:
\begin{equation}
    P_t(\mathbf z)=\operatorname{Norm}\!\left(
    A_{2,t}\tanh(A_{1,t}\mathbf z)\right).
    \label{eq:projector}
\end{equation}
Here $A_{1,t}\in\mathbb R^{h\times m}$ and $A_{2,t}\in\mathbb R^{m\times h}$, with hidden dimension $h$, and $\operatorname{Norm}$ denotes $\ell_2$ normalization; bias terms are omitted for clarity.

The three elicitors are trained separately by minimizing
\begin{equation}
    \mathcal L_t=\mathbb E_d\left[
    \left\|P_t(E_D(d))-E_D(r_t(d))\right\|_2^2\right].
    \label{eq:projector_loss}
\end{equation}
Appendix~\ref{app:projector} provides the teacher prompts and training settings.

\subsection{Shared Sparse Dictionary}
\label{sec:dictionary}

We use an SAE \citep{cunningham2023sparse/sae} to decompose dense representations into sparse features whose meanings can be described in natural language.
For an input $\mathbf x\in\mathbb R^m$, its encoder $g$ and decoder yield
\begin{equation}
    \mathbf c=g(\mathbf x),\qquad
    \hat{\mathbf x}=W\mathbf c+\mathbf b,
    \label{eq:sae}
\end{equation}
where $\mathbf c\in\mathbb R^M$ is a nonnegative sparse code, $W\in\mathbb R^{m\times M}$ is the feature dictionary, and $\mathbf b$ is the decoder bias.
The dictionary contains $M$ features; column $\mathbf w_j$ denotes the direction associated with feature $j$.

We train the SAE on a corpus combining query and document embeddings with embeddings of the teacher-generated texts for the three document perspectives.
Training on these varied semantic expressions encourages the SAE to learn more interpretable features.

We adopt a TopK SAE~\citep{gao2024scaling}, which limits each sparse code to at most $k$ active features.
We learn the dictionary by minimizing reconstruction error:
\begin{equation}
    \mathcal L_{\mathrm{SAE}}=
    \mathbb E_{\mathbf x}\!\left[
    \left\|\mathbf x-(Wg(\mathbf x)+\mathbf b)\right\|_2^2
    \right].
    \label{eq:sae_loss}
\end{equation}

For each feature $j$, we collect highly activating texts and use an LLM to produce a concise description $h_j$, following automated feature-interpretation approaches \citep{paulo2024automatically,park2025decoding}.
The dictionary and descriptions together provide a vocabulary for expressing query--document relations.

\subsection{Retrieval Explanation}
\label{sec:explanation}

Feature interpretation describes the semantics of individual features, while retrieval explanation relates these features to the retriever's score for a query--document pair.
Xetrieval uses joint feature activations to identify semantic connections between the two texts.

We consider the original and elicited representations of a document $d$:
\begin{equation}
    \mathcal V(d)=\{\mathbf z_d\}\cup
    \{P_t(\mathbf z_d):t\in\mathcal T\}.
    \label{eq:views}
\end{equation}
We apply the shared SAE to $\mathbf q$ and each $\mathbf v\in\mathcal V(d)$, selecting features activated by both the query and at least one document representation:
\begin{equation}
    \begin{split}
    \mathcal O(q,d)=\{j:\;&g(\mathbf q)_j>0\ \land\\
    &\exists\,\mathbf v\in\mathcal V(d),\ g(\mathbf v)_j>0\}.
    \end{split}
    \label{eq:overlap}
\end{equation}
The shared features and their natural-language descriptions form the retrieval explanation:
\begin{equation}
    \mathcal E(q,d)=\{(j,h_j):j\in\mathcal O(q,d)\}.
    \label{eq:explanation}
\end{equation}
Each feature $j$ pairs a semantic description $h_j$ with a decoder direction $\mathbf w_j$.

\section{Experiments}
\label{sec:experiments}

We first assess sparse feature quality and interpretability.
We then examine the selected features' influence on relevance scores and their utility for task-level retrieval steering.
These analyses jointly assess the quality of retrieval explanations.
We further examine the use of elicited document views in retrieval.
Finally, we assess the computational efficiency of Xetrieval.

\subsection{Experimental Setting}
\label{sec:exp_setup}

\paragraph{Data and models.}
We evaluate on BRIGHT~\citep{su2024bright}, Natural Questions (NQ; \citealp{kwiatkowski2019natural}), MuTual~\citep{cui2020mutual}, TREC-NEWS~\citep{soboroff2019trec}, Signal-1M~\citep{suarez2018data}, Robust04~\citep{96071}, and ArguAna~\citep{wachsmuth2018retrieval}.
These tasks cover reasoning-intensive retrieval, question answering, dialogue, news, and argument retrieval.
We report NDCG@10 as our retrieval evaluation metric.

The principal explanation configuration uses e5-large~\citep{wang2022text}, DeepSeek-V3~\citep{liu2024deepseek} for semantic supervision, and Qwen3-32B~\citep{yang2025qwen3} for automated feature descriptions and evaluation.
We report retrieval results for six encoders and six teachers, with details provided in Appendix~\ref{app:retrieval}.
We train Feature Elicitors and the shared sparse dictionary using subsets of the StackExchange dataset~\citep{lambert2023huggingface}.
Appendices~\ref{app:projector}--\ref{app:description} give training and feature-evaluation details.

\paragraph{Comparison methods.}
We compare with an SAE baseline for retrieval explanation adapted from the concept-description framework of \citet{park2025decoding}.
It learns sparse features from original embeddings with the same TopK sparsity as Xetrieval.
For retrieval explanation, it selects features active in both the query and document sparse codes.
We refer to this baseline as \emph{SAE} throughout and evaluate it under the same description, intervention, and modulation protocols as Xetrieval.
Following \citet{korznikov2026sanity}, we include an \emph{Untrained SAE} as a control in the description evaluation.

\subsection{Semantic Structure and Feature Interpretability}
\label{sec:semantic_analysis}

\begin{figure*}[t]
  \centering
  \includegraphics[width=\textwidth]{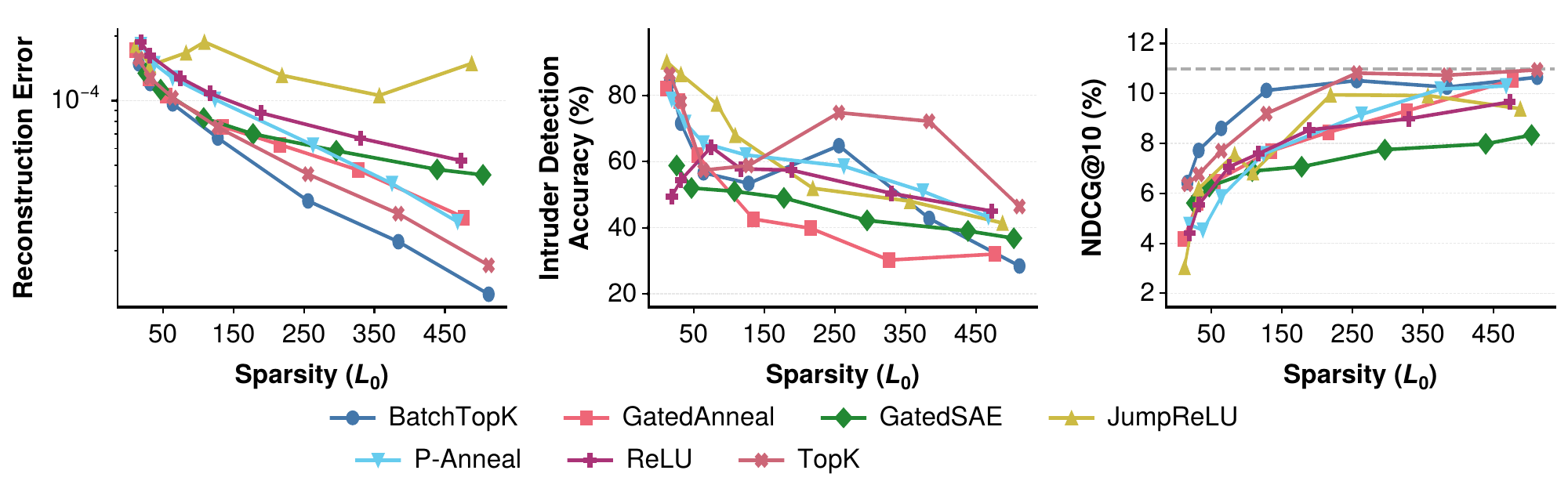}
  \caption{Sparse dictionary quality across sparsity levels. Left: reconstruction error; middle: intruder detection accuracy; right: retrieval with reconstructed embeddings. Reconstruction and retrieval are evaluated on BRIGHT Biology.}
  \label{fig:sae_selection}
\end{figure*}

\paragraph{Sparse dictionary quality.}
A sparse dictionary should preserve retrieval information while yielding semantically coherent features.
We compare seven SAE variants\footnote{Appendix~\ref{app:dictionary} provides SAE training and evaluation details.} using reconstruction error, retrieval with reconstructed embeddings, and intruder detection accuracy, which measures how accurately an LLM identifies a non-activating text among activating texts.
We denote the sparsity setting by $L_0$.
As shown in Figure~\ref{fig:sae_selection}, TopK at $L_0=256$ offers a favorable balance of reconstruction quality, semantic consistency, and retrieval performance after reconstruction.
We therefore adopt this configuration as the default for the subsequent explanation experiments.

\begin{figure}[t]
  \centering
  \includegraphics[width=\columnwidth]{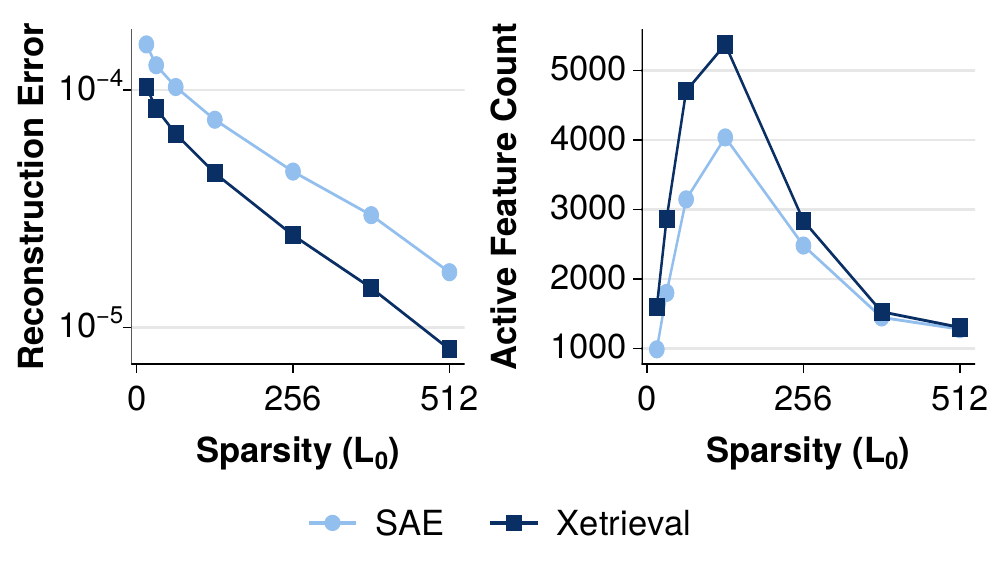}
  \caption{Reconstruction error (left) and distinct active feature counts (right) on BRIGHT for SAE and Xetrieval.}
  \label{fig:semantic_structure}
\end{figure}

\paragraph{Structure of the sparse representations.}
We analyze how feature elicitation affects the sparse representation of document content.
Figure~\ref{fig:semantic_structure} shows that Xetrieval consistently achieves lower reconstruction error than the SAE baseline across sparsity levels from 16 to 512.
We also measure \emph{Active Feature Count}: for the evaluated embedding set $\mathcal X$, this is
\begin{equation}
 N_{\mathrm{active}}(\mathcal X)
 =\left|\bigcup_{\mathbf x\in\mathcal X}\{j:g(\mathbf x)_j>0\}\right|.
\end{equation}
Xetrieval activates more distinct dictionary features than SAE.
This broadens the set of candidate features available for retrieval explanation.
We next examine whether these features also admit consistent semantic descriptions.

\paragraph{Correspondence between descriptions and activations.}
\begin{figure}[t]
  \centering
  \includegraphics[width=\columnwidth]{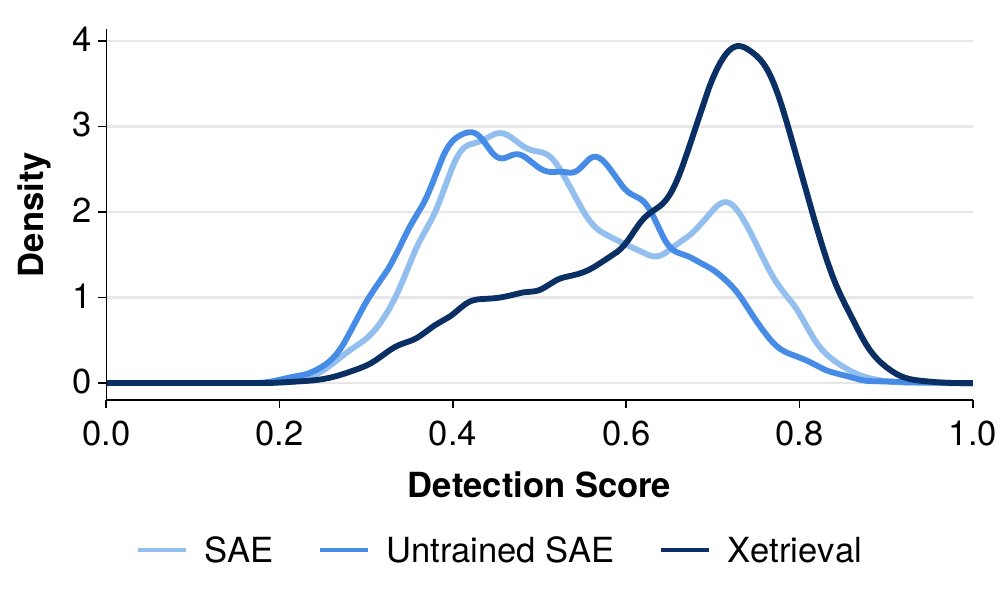}
  \caption{Detection Score distributions for Xetrieval, SAE, and Untrained SAE. Scores measure agreement between feature descriptions and text activation labels.}
  \label{fig:detection}
\end{figure}

We evaluate feature descriptions using the Detection Score of \citet{paulo2024automatically}.
Given a description and a balanced set of activating and non-activating texts, Qwen3-32B predicts which texts exhibit the described feature; classification accuracy is the Detection Score.
Figure~\ref{fig:detection} shows a shift toward higher scores for Xetrieval relative to SAE and Untrained SAE, indicating closer correspondence between its descriptions and the evaluated activation patterns.

\paragraph{Case study.}
\begin{table*}[t]
\centering
\begingroup
\small
\definecolor{CaseElicitation}{RGB}{239,228,244}
\newcommand{\caseactive}{\ensuremath{\bullet}}
\newcommand{\caseinactive}{\ensuremath{\circ}}
\newcommand{\casena}{\textemdash}
\renewcommand{\arraystretch}{1.12}
\begin{tabular}{@{}p{\textwidth}@{}}
\toprule
\textbf{[Query]}~\textit{A question on Marx' ``Value, price and profit'': In his lecture, Karl Marx argues that profit is made by capitalists by selling commodities for their real price, paying workers the real value of commodities they produce but letting them work more time than needed...} \\
\midrule
\textbf{[Document]}~\textit{It is the employing capitalist who immediately extracts from the labourer this surplus value, whatever part of it he may ultimately be able to keep for himself. Upon this relation, therefore between the employing capitalist and the wages labourer the whole wages system and the whole present system of
production hinge. Some of the citizens who took part in our debate were, there, wrong in trying to mince matters, and to treat this fundamental relation between the employing capitalist and the working man as a secondary question, although they were right in stating that, under given circumstances, a rise of prices might affect in very unequal degrees the employing capitalist, the landlord ...} \\
\bottomrule
\end{tabular}

\vspace{5pt}
\setlength{\tabcolsep}{5pt}
\renewcommand{\arraystretch}{1.2}
\begin{tabularx}{\textwidth}{p{0.115\textwidth}>{\raggedright\arraybackslash}X c c c}
\toprule
\textbf{Method} & \textbf{Shared feature description} & \textbf{Query} & \multicolumn{2}{c}{\textbf{Document}} \\
\cmidrule(lr){4-5}
 & & & Original & Elicited \\
\midrule
\textbf{SAE} & Economic theory explanations with conceptual frameworks
 & \caseactive & \caseactive & \casena \\
 & Detailed explanations of concepts
 & \caseactive & \caseactive & \casena \\
\midrule
\textbf{Xetrieval} & Capitalist employment and wage labor
 & \caseactive & \caseactive & \caseactive \\
\rowcolor{CaseElicitation}
 & Surplus value and capitalist profit
 & \caseactive & \caseinactive & \caseactive \\
\rowcolor{CaseElicitation}
 & Prices, wages, and income distribution
 & \caseactive & \caseinactive & \caseactive \\
\bottomrule
\end{tabularx}
\vspace{4pt}
{\footnotesize
\caseactive\ active\qquad \caseinactive\ inactive\qquad \casena\ not applicable.\par}
\caption{Shared semantic features for the Marxian profit example. \emph{Elicited} denotes activation in at least one auxiliary document view; shading marks query overlap identified only through these views.}
\label{tab:case_study_3}
\endgroup
\end{table*}

Table~\ref{tab:case_study_3} examines a query about Marx's account of profit and a passage about surplus value.
The passage relates surplus-value extraction to capitalist employment and discusses its distribution, providing concrete connections to the query's account of profit.
SAE's shared descriptions characterize this content broadly as economic exposition, while Xetrieval expresses these connections through features describing capitalist employment, surplus value, and income distribution.
The latter two are shared with the query through elicited document views.
Feature elicitation thus makes relations expressed in the passage accessible to sparse feature discovery, enabling Xetrieval to provide a more informative explanation of document relevance.
Additional geometry and software examples are provided in Appendix~\ref{app:examples}.

\subsection{Feature Contributions and Retrieval Steering}
\label{sec:behavioral_analysis}

We evaluate the selected features through their influence on relevance scores and their utility for task-level steering with decoded representations.

\subsubsection{Pair-Level Interventions}
\label{sec:interventions}

\begin{figure}[!t]
  \centering
  \includegraphics[width=\columnwidth]{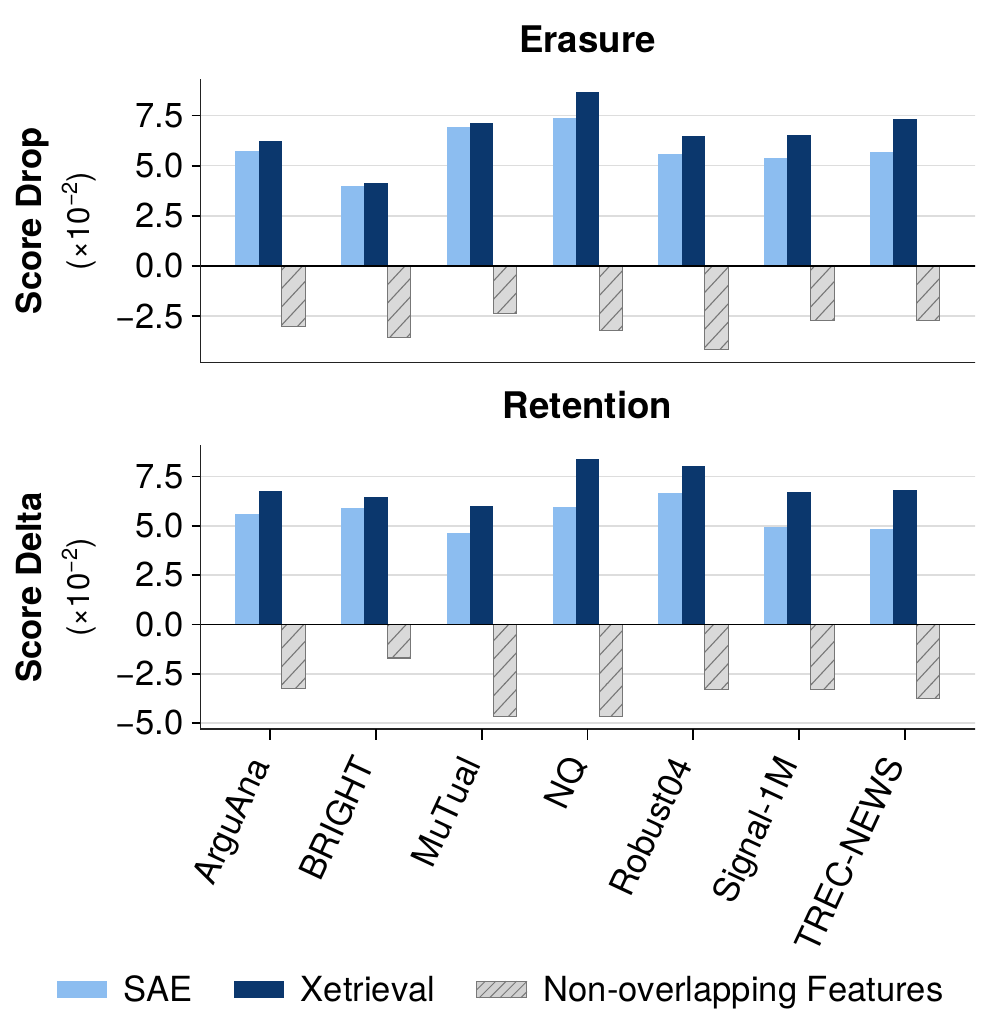}
  \caption{Feature-span interventions on original document embeddings. Positive values indicate a cosine similarity decrease after erasure or increase after retention, relative to the unmodified score. The non-overlapping control uses Xetrieval's dictionary.}
  \label{fig:interventions}
\end{figure}

\begin{figure*}[!t]
  \centering
  \includegraphics[width=\textwidth]{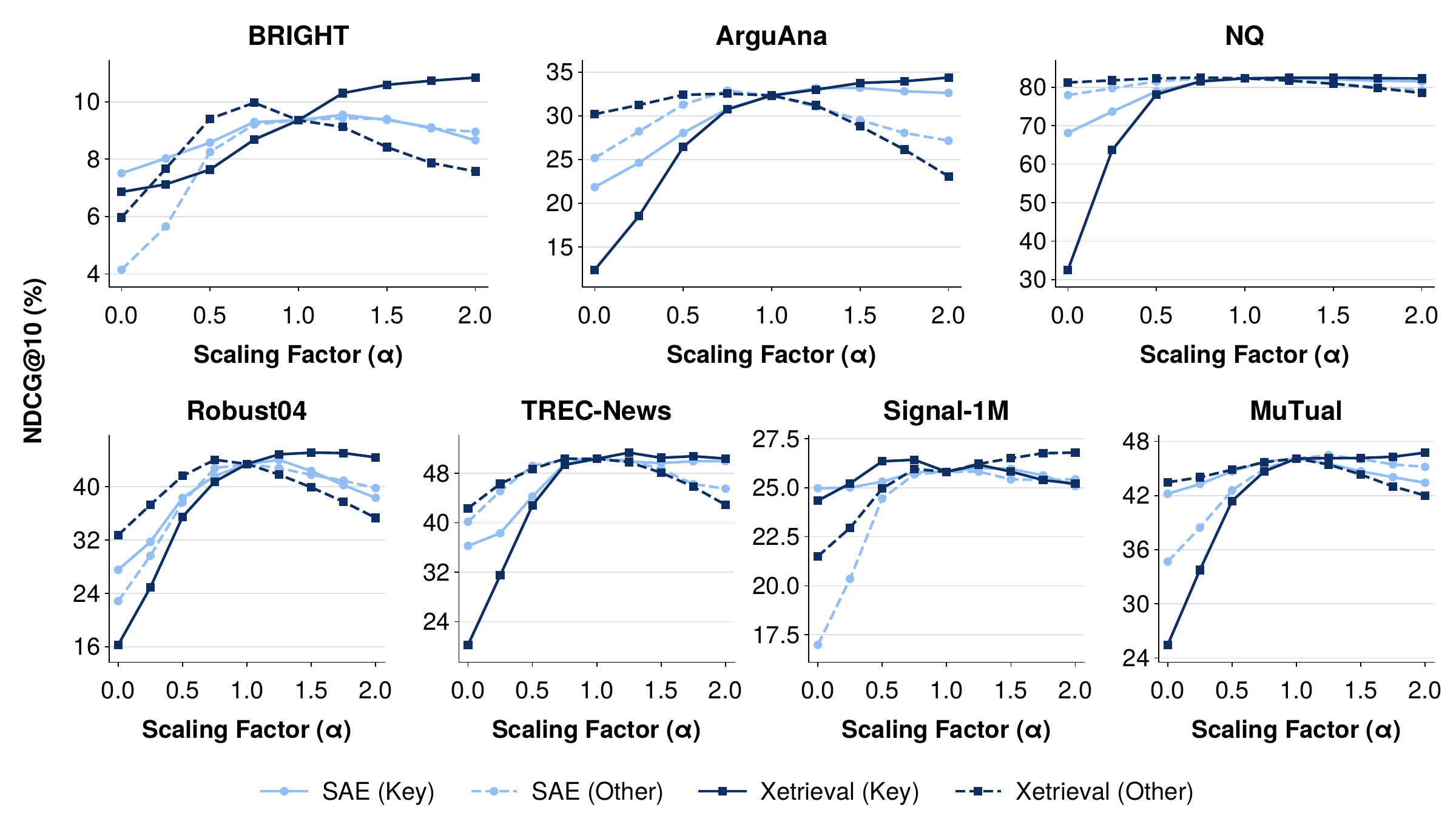}
  \caption{Task-level feature interventions across seven datasets. Features are selected using 30\% of queries and evaluated on the remaining 70\%. Scaling by $\alpha<1$ suppresses the selected activations; $\alpha>1$ amplifies them. The reference $\alpha=1$ uses SAE-decoded representations. The Key and Other feature sets have the same size within each comparison.}
  \label{fig:modulation}
\end{figure*}

To assess whether the selected features contribute to the retriever's relevance score, we intervene on the original document embedding along their decoder directions.
The query embedding remains fixed.
Let $\Pi_S$ be the regularized projection onto the span of decoder directions indexed by a feature set $S$.\footnote{Appendix~\ref{app:intervention} provides feature selection, projection, and sampling details.}
We edit the document embedding by erasure or retention:
\begin{align}
 \mathbf z_d^{\setminus S}
 &=\mathbf z_d-\Pi_S(\mathbf z_d-\mathbf b),\label{eq:erasure}\\
 \mathbf z_d^S
 &=\mathbf b+\Pi_S(\mathbf z_d-\mathbf b).\label{eq:retention}
\end{align}
We normalize the edited embedding and score it against the query using the same cosine function as $s_0$.
To control for feature count, we rank Xetrieval's overlapping features by joint query--document activation and select $S\subseteq\mathcal O(q,d)$ to match the SAE baseline for each pair.
The \emph{Non-overlapping Features} control uses features active in Xetrieval's sparse code for the document but outside $\mathcal O(q,d)$.

As shown in Figure~\ref{fig:interventions}, erasing Xetrieval's feature spans reduces similarity on all seven datasets, with larger average decreases than SAE.
Conversely, retaining the selected span increases similarity, and Xetrieval yields larger increases across the datasets.
Non-overlapping features show the opposite aggregate response under both operations.
Under matched feature counts, Xetrieval thus identifies feature spans with a stronger influence on the original relevance scores than SAE, providing behavioral support for relevance attribution through the selected feature sets.

\subsubsection{Task-Level Interventions}
\label{sec:modulation}

We next examine the utility of these features for task-level retrieval steering.
Within each task, we identify features associated with relevance using the Retrieval Utility Score (RUS), which contrasts co-activation counts in relevant and non-relevant training pairs:
\begin{equation}
 \mathrm{RUS}(j)=
 \sum_{(q,d)\in\mathcal D_{\mathrm{pos}}}I_j(q,d)
 -\sum_{(q,d)\in\mathcal D_{\mathrm{neg}}}I_j(q,d),
 \label{eq:rus}
\end{equation}
Here, $\mathcal D_{\mathrm{pos}}$ and $\mathcal D_{\mathrm{neg}}$ contain relevant and non-relevant training pairs, respectively; $I_j(q,d)$ is one when feature $j$ is active in both the query and document sparse codes, and zero otherwise.

We select the highest-scoring features as the key set and compare it with a same-sized set of other features.
Xetrieval combines the original and elicited document codes, while SAE uses the original document code.
We scale the selected activations in both query and document codes by $\alpha$ before decoding, suppressing features for $\alpha<1$ and amplifying them for $\alpha>1$.
Retrieval is evaluated on test queries by similarity of the decoded representations, with $\alpha=1$ as the reference.
Appendix~\ref{app:intervention} details the query split and modulation procedure.

As shown in Figure~\ref{fig:modulation}, suppressing Xetrieval's key features reduces retrieval performance on most datasets.
Amplifying these features maintains or improves performance, whereas amplifying other features lowers it.
Relative to SAE, Xetrieval generally yields larger losses under key-feature suppression and greater gains under amplification.
These responses demonstrate the utility of Xetrieval's feature selection for task-level retrieval steering.

\begin{table*}[!t]
\centering
\small
\setlength{\tabcolsep}{4pt}
\begin{tabular}{llcccccccc}
\toprule
\textbf{Retriever} & \textbf{Condition} & \textbf{BRIGHT} & \textbf{NQ} & \textbf{MuTual} & \textbf{NEWS} & \textbf{Signal} & \textbf{Robust} & \textbf{ArguAna} & \cellcolor{blue!5!white}{\textbf{Avg.}} \\
\midrule
\multirow{2}{*}{\texttt{e5-small}} & Base & 23.4 & 77.0 & 38.5 & 86.3 & 60.8 & 70.4 & 29.1 & \cellcolor{blue!5!white}{55.1} \\
 & + FE & \textbf{29.6} & \textbf{77.3} & \textbf{38.8} & \textbf{87.5} & \textbf{65.0} & \textbf{75.4} & \textbf{35.7} & \cellcolor{blue!5!white}{\textbf{58.5}} \\
\midrule
\multirow{2}{*}{\texttt{e5-base}} & Base & 30.8 & 81.9 & 40.3 & 90.3 & 68.1 & 76.5 & 32.9 & \cellcolor{blue!5!white}{60.1} \\
 & + FE & \textbf{36.0} & \textbf{82.5} & \textbf{42.1} & \textbf{91.1} & \textbf{71.4} & \textbf{79.0} & \textbf{39.2} & \cellcolor{blue!5!white}{\textbf{63.0}} \\
\midrule
\multirow{2}{*}{\texttt{e5-large}} & Base & 31.5 & 83.3 & \textbf{47.1} & \textbf{90.4} & 66.8 & 77.3 & 34.2 & \cellcolor{blue!5!white}{61.5} \\
 & + FE & \textbf{37.9} & \textbf{84.2} & 46.5 & 90.3 & \textbf{70.3} & \textbf{81.1} & \textbf{39.2} & \cellcolor{blue!5!white}{\textbf{64.2}} \\
\midrule
\multirow{2}{*}{\texttt{gte-base}} & Base & 37.0 & \textbf{81.0} & 28.8 & 92.2 & 73.8 & 77.1 & \textbf{41.7} & \cellcolor{blue!5!white}{61.7} \\
 & + FE & \textbf{39.0} & 80.8 & \textbf{29.6} & \textbf{92.3} & \textbf{74.2} & \textbf{80.2} & 40.9 & \cellcolor{blue!5!white}{\textbf{62.4}} \\
\midrule
\multirow{2}{*}{\texttt{gte-large}} & Base & 41.2 & \textbf{83.0} & 31.2 & 92.0 & \textbf{73.6} & 79.3 & \textbf{41.8} & \cellcolor{blue!5!white}{63.2} \\
 & + FE & \textbf{42.3} & 82.6 & \textbf{31.3} & 92.0 & 72.8 & \textbf{81.6} & 41.3 & \cellcolor{blue!5!white}{\textbf{63.4}} \\
\midrule
\multirow{2}{*}{\texttt{snowflake}} & Base & 34.8 & 48.1 & 36.2 & 22.5 & 64.8 & 24.1 & 37.2 & \cellcolor{blue!5!white}{38.3} \\
 & + FE & \textbf{38.8} & \textbf{68.9} & \textbf{36.3} & \textbf{64.9} & \textbf{67.9} & \textbf{42.7} & \textbf{38.6} & \cellcolor{blue!5!white}{\textbf{51.2}} \\
\bottomrule
\end{tabular}
\caption{NDCG@10 (\%) across six retrievers with DeepSeek-V3 supervision. Base: Base Retriever; + FE: + Feature Elicitation. NEWS, Signal, and Robust denote TREC-NEWS, Signal-1M, and Robust04. Avg. averages the seven tasks. Bold denotes the higher score within each retriever.}
\label{tab:retrieval_main}
\end{table*}

Signal-1M is the only exception to this pattern.
A plausible explanation is that its short, topically diverse tweets~\citep{suarez2018data} make it harder to identify a common set of task-level key features.

Together, these analyses show that Xetrieval yields more interpretable features and selects spans with greater influence on relevance scores.
Task-level modulation further demonstrates the features' utility for retrieval steering.

\subsection{Semantic Presentation in Relevance Matching}
\label{sec:retrieval_effects}

We further investigate how the elicited views affect retrieval performance.
Following \citet{chen2025enrichindex}, we sum the query's similarities to the original and elicited document representations:
\begin{equation}
 s_{\mathrm{aug}}(q,d)=\langle\mathbf q,\mathbf z_d\rangle
 +\sum_{t\in\mathcal T}\langle\mathbf q,P_t(\mathbf z_d)\rangle,
 \label{eq:augmented_score}
\end{equation}
where $\langle\cdot,\cdot\rangle$ denotes cosine similarity.
We compare this condition, denoted \emph{+ Feature Elicitation}, with the \emph{Base Retriever}, which uses the original query--document relevance score.

As shown in Table~\ref{tab:retrieval_main}, Feature Elicitation improves average retrieval performance.
The same trend holds across supervision sources, as detailed in Appendix~\ref{app:retrieval}.
The magnitude of these gains varies across encoders and tasks.
These results suggest that the document views learned for feature discovery can also support more effective retrieval.
The gains are achieved with frozen query and document encoders, indicating that changes in semantic presentation can improve relevance matching.

\subsection{Computational Efficiency}
\label{sec:efficiency}

\begin{figure}[!t]
  \centering
  \includegraphics[width=\columnwidth]{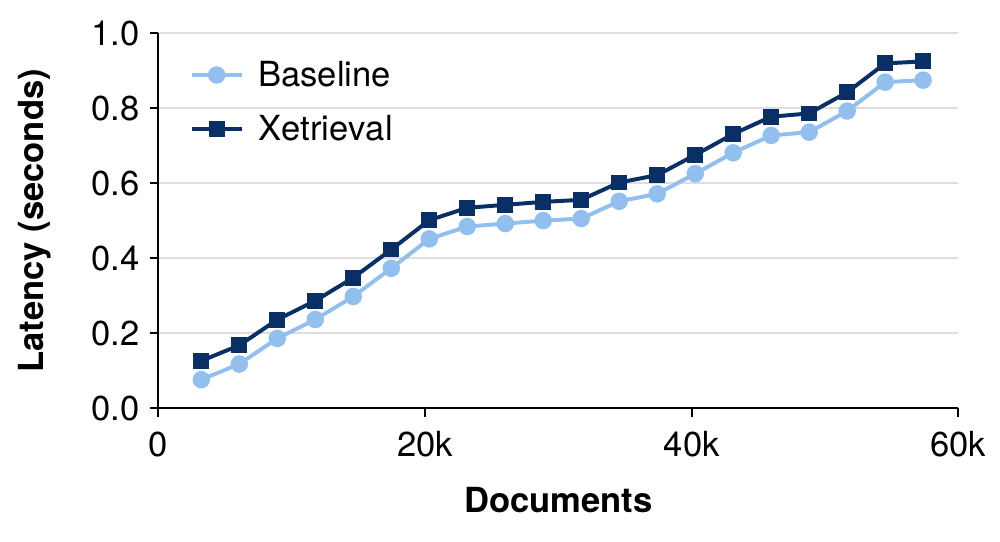}
  \caption{Processing latency of the baseline and Xetrieval on BRIGHT Biology as the number of documents increases. Xetrieval incurs a small absolute latency overhead for feature elicitation and SAE encoding of the auxiliary views.}
  \label{fig:efficiency}
\end{figure}

We assess the computational overhead of Xetrieval using the BRIGHT Biology document set, varying the number of documents on an NVIDIA A100 40GB with end-to-end timing.
As shown in Figure~\ref{fig:efficiency}, Xetrieval incurs a small absolute latency overhead.
First, each Feature Elicitor is an MLP with a single hidden layer, requiring only one forward pass to produce a document view.
Second, sparse decomposition of the additional views requires only feed-forward SAE encoding over fixed-dimensional vectors, which can be processed in a batch.

\section{Conclusion}
\label{sec:conclusion}

We presented Xetrieval, a feature elicitation framework for explaining dense retrieval.
Compared with SAE, Xetrieval yields more interpretable features and identifies feature spans that more strongly influence the retriever's original relevance scores.
Task-level modulation further demonstrates more effective retrieval steering with decoded representations.
Additional analyses show retrieval gains from elicited views and low overhead for elicitation and sparse encoding.
These findings highlight how semantic presentation shapes the interpretable features available for relevance attribution.

\FloatBarrier
\section*{Limitations}

Xetrieval requires access to query and document embeddings and cannot be directly applied to retrieval services that expose only rankings or relevance scores.
SAE dictionaries depend on the corpora used for training, and adapting Xetrieval to new languages or specialized domains may require retraining its feature elicitors and sparse dictionary on representative corpora.
Our feature descriptions and interpretability assessments rely on LLMs, whose outputs can vary with model choice and prompting.

\section*{Ethical Considerations}

This study uses existing text corpora and pretrained models to explain dense retrieval.
Teacher-generated texts and feature descriptions can reproduce biases, omit context, or assign an overly broad semantic label to an activation pattern.
Explanations should therefore be read alongside the source query and document, especially when they inform decisions in a specialized domain.
Their role is to support understanding of the retriever's scores; the factual reliability and relevance of retrieved content require assessment from the source texts.

\paragraph{Artifacts and licenses.}
We use existing datasets, pretrained models, and third-party software in accordance with their respective licenses and terms of use.

\paragraph{LLM assistance.}
LLMs were used for the teacher texts, feature descriptions, and automated evaluations described in the methodology.
Generative AI tools also assisted with manuscript organization, drafting, editing, literature lookup, and figure preparation.

\bibliography{custom}
\clearpage
\appendix
\section{Feature Elicitor Training}
\label{app:projector}

\paragraph{Training corpus and semantic supervision.}
We construct paired representations from 11,796 documents drawn from the StackExchange preference dataset~\citep{lambert2023huggingface}.
Table~\ref{tab:lr_domain_distribution} gives the community distribution.
For each document, an LLM generates a Summary, Purpose, and QA perspective using the prompts in Table~\ref{tab:lr_prompts}.
The document and each generated text are embedded by the same retriever.
The QA output, which contains at most 20 question--answer pairs, is encoded as a single text block.
Each perspective defines a separate regression target for its Feature Elicitor.

\begin{table}[t]
\centering
\small
\begin{tabular}{@{}l c l c@{}}
\toprule
\textbf{Community} & \textbf{\# Docs} & \textbf{Community} & \textbf{\# Docs} \\
\midrule
\textit{politics} & 1{,}000 & \textit{mathematica} & 1{,}000 \\
\textit{codereview} & 600 & \textit{economics} & 600 \\
\textit{cs} & 600 & \textit{chemistry} & 600 \\
\textit{StackOverflow} & 600 & \textit{ai} & 600 \\
\textit{bioinformatics} & 600 & \textit{codegolf} & 600 \\
\textit{math} & 600 & \textit{robotics} & 600 \\
\textit{earthscience} & 600 & \textit{mathoverflow} & 600 \\
\textit{biology} & 600 & \textit{philosophy} & 600 \\
\textit{software\-engineering} & 600 & \textit{sustainability} & 432 \\
\textit{computergraphics} & 364 &  &  \\
\midrule
\textbf{Total} & \textbf{11{,}796} &  &  \\
\bottomrule
\end{tabular}
\caption{Domain distribution of the StackExchange corpus used to construct Feature Elicitor training pairs.}
\label{tab:lr_domain_distribution}
\end{table}

\begin{table*}[t]
\centering
\small
\setlength{\tabcolsep}{4pt}
\begin{tabular}{@{}p{0.16\textwidth} >{\raggedright\arraybackslash}p{0.78\textwidth}@{}}
\toprule
\textbf{Type} & \textbf{Prompt} \\
\midrule
\textsc{Purpose} &
{\ttfamily Given the following text, describe the purpose of this text in layman’s terms in one paragraph.\newline\newline\{doc\}\newline} \\
\textsc{Summary} &
{\ttfamily Given the following text, summarize this text in layman’s terms in one paragraph.\newline\newline\{doc\}\newline} \\
\textsc{QA} &
{\ttfamily Given the following text, generate at most 20 distinct question-answer pairs on this text. The questions should be general, and phrased in layman’s terms, using vocabulary that can be distinct from the text, but still requires explicit or implicit knowledge from the text. Only output the question-answer pairs, no other explanation.\newline\newline\{doc\}\newline} \\
\bottomrule
\end{tabular}
\caption{Teacher prompts used to generate the Summary, Purpose, and QA perspectives.}
\label{tab:lr_prompts}
\end{table*}

\paragraph{Architecture and optimization.}
Each elicitor follows Equation~\ref{eq:projector}, with hidden width $h=512$, a $\tanh$ activation, and an $\ell_2$-normalized output.
The three elicitors are optimized separately with mean squared error against the teacher-text embeddings.
We use Adam with learning rate $5\times10^{-4}$, batch size 128, and at most 100 epochs.
The paired embeddings are split into 85\% training and 15\% validation examples; early stopping uses validation loss with patience five.
Training typically takes one to two minutes per elicitor.
Together with the low inference overhead reported in Section~\ref{sec:efficiency}, this shows that Feature Elicitors are efficient both to train and to apply.

\section{Sparse Dictionary Learning and Selection}
\label{app:dictionary}

\paragraph{Corpus and dictionary learning.}
The dictionary-learning corpus expands the elicitor-training collection with additional StackExchange domains and contains 84,860 documents.
DeepSeek-V3 generates the three semantic perspectives using Table~\ref{tab:lr_prompts}.
The SAE is trained on query and document embeddings together with embeddings of these teacher-generated texts.

We compare implementations from \texttt{dictionary\_learning}~\citep{marks2024dictionary_learning}: ReLU~\citep{cunningham2023sparse/sae}, TopK~\citep{gao2024scaling}, BatchTopK~\citep{bussmann2024batchtopk}, Gated~\citep{rajamanoharan2024improving}, JumpReLU~\citep{rajamanoharan2024jumping}, P-Annealing~\citep{karvonen2024measuring}, and GatedAnnealing.
We assess dictionary quality through reconstruction, semantic consistency, and retrieval with reconstructed embeddings~\citep{park2025decoding,karvonen2025saebench}.

\paragraph{Selection criteria.}
Reconstruction Error is the mean squared error between input embeddings and SAE reconstructions, evaluated on BRIGHT Biology documents.
Following \citet{park2025decoding}, we assess semantic consistency through \emph{Intruder Detection Accuracy}, which measures whether an LLM can distinguish a non-activating document from nine highly activating documents for a feature.
Using activations on the SAE training corpus, we uniformly sample 500 eligible dictionary features; for each, we present its nine highest-activation documents and one randomly sampled non-activating document to Qwen3-32B.
The prompt is given in Table~\ref{tab:ms_prompt}.
Finally, we evaluate NDCG@10 using SAE-reconstructed embeddings on BRIGHT Biology.

Figure~\ref{fig:sae_selection} in Section~\ref{sec:semantic_analysis} reports the comparison across sparsity levels.

\begin{table}[t]
\centering
\small
\begin{tabular}{@{}>{\raggedright\arraybackslash}p{\columnwidth}@{}}
\toprule
\textbf{Intruder Detection Prompt}\\
\midrule
{\ttfamily
You are an expert linguist analyzing pieces of documents. Below, you will see a set of documents that has some common features, but one of them is an intruder (it does not have that common feature in it).\newline
Your task is to identify the intruder document and explain why it does not fit.\newline\newline
The last line of your response must be the formatted response, using \newline``[intruder]:Document\#''\newline\newline
\{documents\}\newline\newline
Which document is the intruder, and why?}\\
\bottomrule
\end{tabular}
\caption{Prompt used for Intruder Detection Accuracy evaluation.}
\label{tab:ms_prompt}
\end{table}

\section{Feature Description and Evaluation}
\label{app:description}

Following \citet{paulo2024automatically}, we generate natural-language feature descriptions from activating texts and assess their agreement with activation patterns using Detection.
For each feature, Qwen3-32B receives highly activating texts and produces a concise natural-language description.
The prompt requests a description of one to eight words and provides the activation-associated importance of each example.
These descriptions serve as the semantic labels of features in query--document explanations.
Table~\ref{tab:explain_eval_prompts} gives both the description and Detection prompts.

Detection evaluates the agreement between feature descriptions and activation patterns through description-conditioned classification.
The evaluator receives a feature description and balanced activating and non-activating texts and predicts which texts exhibit the described feature.
We compute classification accuracy separately for each feature and plot its distribution using kernel density estimation.

\begin{table*}[t]
\centering
\small
\renewcommand{\arraystretch}{1.15}
\begin{tabular}{@{}p{0.10\textwidth} p{\dimexpr0.90\textwidth-2\tabcolsep\relax}@{}}
\toprule
\textbf{Prompt} & \textbf{Content} \\
\midrule
\textbf{Describe} &
\begin{minipage}[t]{\linewidth}
\raggedright
\ttfamily
You are a meticulous AI researcher conducting an important investigation into patterns found in language. Your task is to analyze text and provide an interpretation that thoroughly encapsulates possible patterns found in it.\par
\par
Guidelines: You will be given a list of text examples on which a certain common pattern might be present. How important each text is for the pattern is listed after each text.\par
\par
- Try to produce a concise final description. Simply describe the text latents that are common in the examples, and what patterns you found.\par
- If the examples are uninformative, you don’t need to mention them. Don’t focus on giving examples of important tokens, but try to summarize the patterns found in the examples.\par
- Based on the found patterns, summarize your interpretation in 1--8 words.\par
- Do not make lists of possible interpretations. Keep your interpretations short and concise.\par
- The last line of your response must be the formatted interpretation, using [EXPLANATION]:
\end{minipage}
\\
\midrule
\textbf{Evaluate} &
\begin{minipage}[t]{\linewidth}
\raggedright
\ttfamily
You are an intelligent and meticulous linguistics researcher.\par
\par
You will be given a latent explanation (a hypothesis) that describes a sentence-level concept.\par
\par
You will then be given several full text examples (each is a whole sentence/document). Your task is to determine which examples possess the latent implied by the explanation.
\end{minipage}
\\
\bottomrule
\end{tabular}
\caption{Prompts used to generate feature descriptions and evaluate Detection Scores.}
\label{tab:explain_eval_prompts}
\end{table*}

\section{Intervention and Modulation Details}
\label{app:intervention}

\subsection{Feature-Span Interventions}

\paragraph{Pair sampling.}
We rank documents by the cosine similarity of original query and document embeddings.
For each query, the top 32 documents form the candidate pool.
This pool includes documents annotated as relevant and retrieved documents without a relevance annotation.
Duplicate documents are removed within each query, and at most four pairs are sampled from the combined pool using a fixed random seed.
The query remains unchanged in every intervention.

\paragraph{Feature sets.}
With $\mathrm{supp}_{+}(\mathbf c)=\{j:c_j>0\}$, SAE selects
\begin{equation}
 \begin{split}
 S_{\mathrm{SAE}}(q,d)={}&
 \mathrm{supp}_{+}(g_{\mathrm{SAE}}(\mathbf q))\\
 &\cap\mathrm{supp}_{+}(g_{\mathrm{SAE}}(\mathbf z_d)),
 \end{split}
\end{equation}
where $g_{\mathrm{SAE}}$ is the SAE trained on original embeddings.
To control for feature count in pair-level interventions, we match the number of selected features between Xetrieval and SAE for each query--document pair.
Let $n=|S_{\mathrm{SAE}}(q,d)|$.
For each feature $j\in\mathcal O(q,d)$, we compute its joint activation score:
\begin{equation}
    \omega_j(q,d)
    = g(\mathbf q)_j
      \sum_{\mathbf v\in\mathcal V(d)} g(\mathbf v)_j,
    \label{eq:joint_activation}
\end{equation}
where $\mathcal V(d)$ contains the original document embedding and its three elicited views.
Xetrieval selects the $n$ features with the highest scores from $\mathcal O(q,d)$.
The selected decoder directions define the span used for erasure and retention.
The non-overlapping control uses features active in the  document under Xetrieval's dictionary, excluding the full overlap $\mathcal O(q,d)$.

\paragraph{Regularized span projection.}
A zero SAE coefficient does not imply that the centered document embedding is orthogonal to the corresponding decoder direction.
For selected decoder columns $W_S$, the intervention operator is
\begin{equation}
 \Pi_S(\mathbf u)=W_S(W_S^\top W_S+\lambda I)^{-1}W_S^\top\mathbf u,
\end{equation}
where $I$ is the $|S|\times|S|$ identity matrix and $\lambda=10^{-6}$ is the regularization coefficient.
We apply this operator to the original document embedding centered by the decoder bias: $\Pi_S(\mathbf z_d-\mathbf b)$ extracts its component along the selected directions.
Equations~\ref{eq:erasure} and~\ref{eq:retention} then remove or retain this component.
The resulting embeddings are normalized before scoring with the original query.
Erasure is reported as $\cos(\mathbf q,\mathbf z_d)-\cos(\mathbf q,\mathbf z_d^{\setminus S})$; retention is reported as $\cos(\mathbf q,\mathbf z_d^S)-\cos(\mathbf q,\mathbf z_d)$.
Figure~\ref{fig:interventions} averages these quantities over the sampled pairs of each dataset.
These interventions evaluate each selected feature span as a whole in the base retriever's representation space.

\subsection{Feature Modulation}

The RUS calculation in Equation~\ref{eq:rus} uses the training queries from a 30\%/70\% query split, and ranking is evaluated on the remaining test queries.
All pairs belonging to a query stay within its assigned split.
For Xetrieval, RUS counts a feature as co-active when it is active in the query and at least one of the four document representations; for SAE, co-activation uses only the original query and document representations.
We select 1,024 key features on ArguAna, NQ, and Signal-1M, and 256 on BRIGHT, Robust04, TREC-NEWS, and MuTual.
The comparison set contains the same number of other features, chosen from outside the key set.
For a selected set $S$, modulation changes both query and document sparse codes according to
\begin{equation}
 c^{(\alpha,S)}_j=\begin{cases}
 \alpha c_j,&j\in S,\\
 c_j,&j\notin S.
 \end{cases}
\end{equation}
We decode the modified codes and evaluate retrieval as $\alpha$ ranges from zero to two.

\section{Complete Retrieval Results}
\label{app:retrieval}

\paragraph{Encoders and teachers.}
We report results for six encoders: e5-small, e5-base, and e5-large~\citep{wang2022text}; gte-base and gte-large~\citep{li2023towards}; and Snowflake-Arctic-Embed~\citep{yu2024arctic}.
We use six teachers to generate supervision: DeepSeek-V2-Lite~\citep{liu2024deepseek-v2}, DeepSeek-V3~\citep{liu2024deepseek}, DeepSeek-R1~\citep{guo2025deepseek-r1}, Qwen3-32B~\citep{yang2025qwen3}, and GPT-OSS-20B and GPT-OSS-120B~\citep{agarwal2025gpt}.
Feature Elicitors are learned separately for each encoder's representation space and each teacher's targets.

\paragraph{Results across supervision sources.}
Tables~\ref{tab:lr_result_r1}--\ref{tab:lr_result_qwen3} report the encoder--teacher--task comparisons.
The Base Retriever uses the original query and document embeddings.
Feature Elicitation uses Equation~\ref{eq:augmented_score} with the elicited document representations.
Across all six teachers, Feature Elicitation improves average retrieval performance for each evaluated encoder.

\begin{table*}[p]
\centering
\small
\setlength{\tabcolsep}{4pt}
\begin{tabular}{llcccccccc}
\toprule
\textbf{Retriever} & \textbf{Condition} & \textbf{BRIGHT} & \textbf{NQ} & \textbf{MuTual} & \textbf{NEWS} & \textbf{Signal} & \textbf{Robust} & \textbf{ArguAna} & \cellcolor{blue!5!white}{\textbf{Avg.}} \\
\midrule
\multirow{2}{*}{\texttt{e5-base}} & Base & 30.8 & 81.9 & 40.3 & 90.3 & 68.1 & 76.5 & 32.9 & \cellcolor{blue!5!white}{60.1} \\
 & + FE & \textbf{36.8} & \textbf{82.4} & \textbf{41.6} & \textbf{90.5} & \textbf{71.4} & \textbf{79.5} & \textbf{39.8} & \cellcolor{blue!5!white}{\textbf{63.2}} \\
\midrule
\multirow{2}{*}{\texttt{e5-large}} & Base & 31.5 & 83.3 & \textbf{47.1} & \textbf{90.4} & 66.8 & 77.3 & 34.2 & \cellcolor{blue!5!white}{61.5} \\
 & + FE & \textbf{38.7} & \textbf{84.0} & 46.8 & 89.7 & \textbf{69.3} & \textbf{81.8} & \textbf{40.6} & \cellcolor{blue!5!white}{\textbf{64.4}} \\
\midrule
\multirow{2}{*}{\texttt{e5-small}} & Base & 23.4 & 77.0 & 38.5 & 86.3 & 60.8 & 70.4 & 29.1 & \cellcolor{blue!5!white}{55.1} \\
 & + FE & \textbf{30.6} & 77.0 & \textbf{38.8} & \textbf{88.3} & \textbf{64.9} & \textbf{76.0} & \textbf{36.8} & \cellcolor{blue!5!white}{\textbf{58.9}} \\
\midrule
\multirow{2}{*}{\texttt{gte-base}} & Base & 37.0 & \textbf{81.0} & 28.8 & \textbf{92.2} & 73.8 & 77.1 & \textbf{41.7} & \cellcolor{blue!5!white}{61.7} \\
 & + FE & \textbf{39.4} & 80.6 & \textbf{29.9} & 91.8 & \textbf{73.9} & \textbf{80.0} & 40.6 & \cellcolor{blue!5!white}{\textbf{62.3}} \\
\midrule
\multirow{2}{*}{\texttt{gte-large}} & Base & 41.2 & \textbf{83.0} & 31.2 & 92.0 & \textbf{73.6} & 79.3 & \textbf{41.8} & \cellcolor{blue!5!white}{63.2} \\
 & + FE & \textbf{42.8} & 82.8 & \textbf{31.4} & \textbf{92.7} & 73.0 & \textbf{81.3} & 41.5 & \cellcolor{blue!5!white}{\textbf{63.6}} \\
\midrule
\multirow{2}{*}{\texttt{snowflake}} & Base & 34.8 & 48.1 & 36.2 & 22.5 & 64.8 & 24.1 & 37.2 & \cellcolor{blue!5!white}{38.3} \\
 & + FE & \textbf{39.7} & \textbf{69.8} & \textbf{36.9} & \textbf{66.4} & \textbf{68.3} & \textbf{42.9} & \textbf{38.8} & \cellcolor{blue!5!white}{\textbf{51.8}} \\
\bottomrule
\end{tabular}
\caption{NDCG@10 (\%) results with DeepSeek-R1 supervision. Base: Base Retriever; + FE: + Feature Elicitation. NEWS, Signal, and Robust denote TREC-NEWS, Signal-1M, and Robust04. Bold denotes the higher score within each retriever.}
\label{tab:lr_result_r1}
\end{table*}

\begin{table*}[p]
\centering
\small
\setlength{\tabcolsep}{4pt}
\begin{tabular}{llcccccccc}
\toprule
\textbf{Retriever} & \textbf{Condition} & \textbf{BRIGHT} & \textbf{NQ} & \textbf{MuTual} & \textbf{NEWS} & \textbf{Signal} & \textbf{Robust} & \textbf{ArguAna} & \cellcolor{blue!5!white}{\textbf{Avg.}} \\
\midrule
\multirow{2}{*}{\texttt{e5-base}} & Base & 30.8 & 81.9 & 40.3 & 90.3 & 68.1 & 76.5 & 32.9 & \cellcolor{blue!5!white}{60.1} \\
 & + FE & \textbf{36.0} & \textbf{82.5} & \textbf{42.1} & \textbf{91.1} & \textbf{71.4} & \textbf{79.0} & \textbf{39.2} & \cellcolor{blue!5!white}{\textbf{63.0}} \\
\midrule
\multirow{2}{*}{\texttt{e5-large}} & Base & 31.5 & 83.3 & \textbf{47.1} & \textbf{90.4} & 66.8 & 77.3 & 34.2 & \cellcolor{blue!5!white}{61.5} \\
 & + FE & \textbf{37.9} & \textbf{84.2} & 46.5 & 90.3 & \textbf{70.3} & \textbf{81.1} & \textbf{39.2} & \cellcolor{blue!5!white}{\textbf{64.2}} \\
\midrule
\multirow{2}{*}{\texttt{e5-small}} & Base & 23.4 & 77.0 & 38.5 & 86.3 & 60.8 & 70.4 & 29.1 & \cellcolor{blue!5!white}{55.1} \\
 & + FE & \textbf{29.6} & \textbf{77.3} & \textbf{38.8} & \textbf{87.5} & \textbf{65.0} & \textbf{75.4} & \textbf{35.7} & \cellcolor{blue!5!white}{\textbf{58.5}} \\
\midrule
\multirow{2}{*}{\texttt{gte-base}} & Base & 37.0 & \textbf{81.0} & 28.8 & 92.2 & 73.8 & 77.1 & \textbf{41.7} & \cellcolor{blue!5!white}{61.7} \\
 & + FE & \textbf{39.0} & 80.8 & \textbf{29.6} & \textbf{92.3} & \textbf{74.2} & \textbf{80.2} & 40.9 & \cellcolor{blue!5!white}{\textbf{62.4}} \\
\midrule
\multirow{2}{*}{\texttt{gte-large}} & Base & 41.2 & \textbf{83.0} & 31.2 & 92.0 & \textbf{73.6} & 79.3 & \textbf{41.8} & \cellcolor{blue!5!white}{63.2} \\
 & + FE & \textbf{42.3} & 82.6 & \textbf{31.3} & 92.0 & 72.8 & \textbf{81.6} & 41.3 & \cellcolor{blue!5!white}{\textbf{63.4}} \\
\midrule
\multirow{2}{*}{\texttt{snowflake}} & Base & 34.8 & 48.1 & 36.2 & 22.5 & 64.8 & 24.1 & 37.2 & \cellcolor{blue!5!white}{38.3} \\
 & + FE & \textbf{38.8} & \textbf{68.9} & \textbf{36.3} & \textbf{64.9} & \textbf{67.9} & \textbf{42.7} & \textbf{38.6} & \cellcolor{blue!5!white}{\textbf{51.2}} \\
\bottomrule
\end{tabular}
\caption{NDCG@10 (\%) results with DeepSeek-V3 supervision. Base: Base Retriever; + FE: + Feature Elicitation. NEWS, Signal, and Robust denote TREC-NEWS, Signal-1M, and Robust04. Bold denotes the higher score within each retriever.}
\label{tab:lr_result_v3_appendix}
\end{table*}

\begin{table*}[p]
\centering
\small
\setlength{\tabcolsep}{4pt}
\begin{tabular}{llcccccccc}
\toprule
\textbf{Retriever} & \textbf{Condition} & \textbf{BRIGHT} & \textbf{NQ} & \textbf{MuTual} & \textbf{NEWS} & \textbf{Signal} & \textbf{Robust} & \textbf{ArguAna} & \cellcolor{blue!5!white}{\textbf{Avg.}} \\
\midrule
\multirow{2}{*}{\texttt{e5-base}} & Base & 30.8 & 81.9 & 40.3 & 90.3 & 68.1 & 76.5 & 32.9 & \cellcolor{blue!5!white}{60.1} \\
 & + FE & \textbf{35.3} & \textbf{82.1} & \textbf{41.7} & \textbf{91.0} & \textbf{71.4} & \textbf{79.3} & \textbf{39.8} & \cellcolor{blue!5!white}{\textbf{62.9}} \\
\midrule
\multirow{2}{*}{\texttt{e5-large}} & Base & 31.5 & 83.3 & \textbf{47.1} & 90.4 & 66.8 & 77.3 & 34.2 & \cellcolor{blue!5!white}{61.5} \\
 & + FE & \textbf{37.5} & \textbf{83.9} & 46.7 & \textbf{91.1} & \textbf{71.7} & \textbf{81.7} & \textbf{40.7} & \cellcolor{blue!5!white}{\textbf{64.8}} \\
\midrule
\multirow{2}{*}{\texttt{e5-small}} & Base & 23.4 & 77.0 & 38.5 & 86.3 & 60.8 & 70.4 & 29.1 & \cellcolor{blue!5!white}{55.1} \\
 & + FE & \textbf{27.4} & 77.0 & 38.5 & \textbf{87.8} & \textbf{65.8} & \textbf{75.0} & \textbf{37.3} & \cellcolor{blue!5!white}{\textbf{58.4}} \\
\midrule
\multirow{2}{*}{\texttt{gte-base}} & Base & 37.0 & \textbf{81.0} & 28.8 & \textbf{92.2} & 73.8 & 77.1 & \textbf{41.7} & \cellcolor{blue!5!white}{61.7} \\
 & + FE & \textbf{39.1} & 80.9 & \textbf{29.8} & 91.7 & \textbf{74.1} & \textbf{80.0} & 41.3 & \cellcolor{blue!5!white}{\textbf{62.4}} \\
\midrule
\multirow{2}{*}{\texttt{gte-large}} & Base & 41.2 & \textbf{83.0} & 31.2 & 92.0 & 73.6 & 79.3 & \textbf{41.8} & \cellcolor{blue!5!white}{63.2} \\
 & + FE & \textbf{42.4} & 82.7 & \textbf{31.3} & \textbf{92.3} & \textbf{73.7} & \textbf{81.7} & 41.7 & \cellcolor{blue!5!white}{\textbf{63.7}} \\
\midrule
\multirow{2}{*}{\texttt{snowflake}} & Base & 34.8 & 48.1 & 36.2 & 22.5 & 64.8 & 24.1 & 37.2 & \cellcolor{blue!5!white}{38.3} \\
 & + FE & \textbf{38.3} & \textbf{68.5} & \textbf{37.0} & \textbf{62.2} & \textbf{68.0} & \textbf{40.8} & \textbf{38.7} & \cellcolor{blue!5!white}{\textbf{50.5}} \\
\bottomrule
\end{tabular}
\caption{NDCG@10 (\%) results with DeepSeek-V2-Lite supervision. Base: Base Retriever; + FE: + Feature Elicitation. NEWS, Signal, and Robust denote TREC-NEWS, Signal-1M, and Robust04. Bold denotes the higher score within each retriever.}
\label{tab:lr_result_v2}
\end{table*}

\begin{table*}[p]
\centering
\small
\setlength{\tabcolsep}{4pt}
\begin{tabular}{llcccccccc}
\toprule
\textbf{Retriever} & \textbf{Condition} & \textbf{BRIGHT} & \textbf{NQ} & \textbf{MuTual} & \textbf{NEWS} & \textbf{Signal} & \textbf{Robust} & \textbf{ArguAna} & \cellcolor{blue!5!white}{\textbf{Avg.}} \\
\midrule
\multirow{2}{*}{\texttt{e5-base}} & Base & 30.8 & 81.9 & 40.3 & 90.3 & 68.1 & 76.5 & 32.9 & \cellcolor{blue!5!white}{60.1} \\
 & + FE & \textbf{36.3} & \textbf{82.4} & \textbf{41.5} & \textbf{90.6} & \textbf{71.1} & \textbf{79.6} & \textbf{39.4} & \cellcolor{blue!5!white}{\textbf{63.0}} \\
\midrule
\multirow{2}{*}{\texttt{e5-large}} & Base & 31.5 & 83.3 & \textbf{47.1} & \textbf{90.4} & 66.8 & 77.3 & 34.2 & \cellcolor{blue!5!white}{61.5} \\
 & + FE & \textbf{38.2} & \textbf{84.0} & 46.8 & 89.9 & \textbf{69.5} & \textbf{81.9} & \textbf{40.9} & \cellcolor{blue!5!white}{\textbf{64.5}} \\
\midrule
\multirow{2}{*}{\texttt{e5-small}} & Base & 23.4 & 77.0 & 38.5 & 86.3 & 60.8 & 70.4 & 29.1 & \cellcolor{blue!5!white}{55.1} \\
 & + FE & \textbf{30.4} & 77.0 & \textbf{39.4} & \textbf{88.3} & \textbf{64.6} & \textbf{75.4} & \textbf{35.8} & \cellcolor{blue!5!white}{\textbf{58.7}} \\
\midrule
\multirow{2}{*}{\texttt{gte-base}} & Base & 37.0 & \textbf{81.0} & 28.8 & 92.2 & 73.8 & 77.1 & \textbf{41.7} & \cellcolor{blue!5!white}{61.7} \\
 & + FE & \textbf{39.2} & 80.8 & \textbf{29.8} & 92.2 & 73.8 & \textbf{80.3} & 40.4 & \cellcolor{blue!5!white}{\textbf{62.4}} \\
\midrule
\multirow{2}{*}{\texttt{gte-large}} & Base & 41.2 & \textbf{83.0} & 31.2 & 92.0 & \textbf{73.6} & 79.3 & \textbf{41.8} & \cellcolor{blue!5!white}{63.2} \\
 & + FE & \textbf{42.4} & 82.7 & \textbf{31.7} & \textbf{92.9} & 73.0 & \textbf{81.1} & 40.8 & \cellcolor{blue!5!white}{\textbf{63.5}} \\
\midrule
\multirow{2}{*}{\texttt{snowflake}} & Base & 34.8 & 48.1 & 36.2 & 22.5 & 64.8 & 24.1 & 37.2 & \cellcolor{blue!5!white}{38.3} \\
 & + FE & \textbf{39.2} & \textbf{69.4} & \textbf{36.9} & \textbf{67.0} & \textbf{67.8} & \textbf{43.2} & \textbf{38.8} & \cellcolor{blue!5!white}{\textbf{51.8}} \\
\bottomrule
\end{tabular}
\caption{NDCG@10 (\%) results with GPT-OSS-120B supervision. Base: Base Retriever; + FE: + Feature Elicitation. NEWS, Signal, and Robust denote TREC-NEWS, Signal-1M, and Robust04. Bold denotes the higher score within each retriever.}
\label{tab:lr_result_gpt}
\end{table*}

\begin{table*}[p]
\centering
\small
\setlength{\tabcolsep}{4pt}
\begin{tabular}{llcccccccc}
\toprule
\textbf{Retriever} & \textbf{Condition} & \textbf{BRIGHT} & \textbf{NQ} & \textbf{MuTual} & \textbf{NEWS} & \textbf{Signal} & \textbf{Robust} & \textbf{ArguAna} & \cellcolor{blue!5!white}{\textbf{Avg.}} \\
\midrule
\multirow{2}{*}{\texttt{e5-base}} & Base & 30.8 & 81.9 & 40.3 & 90.3 & 68.1 & 76.5 & 32.9 & \cellcolor{blue!5!white}{60.1} \\
 & + FE & \textbf{36.2} & \textbf{82.4} & \textbf{42.0} & \textbf{90.4} & \textbf{71.5} & \textbf{79.6} & \textbf{39.1} & \cellcolor{blue!5!white}{\textbf{63.0}} \\
\midrule
\multirow{2}{*}{\texttt{e5-large}} & Base & 31.5 & 83.3 & 47.1 & 90.4 & 66.8 & 77.3 & 34.2 & \cellcolor{blue!5!white}{61.5} \\
 & + FE & \textbf{38.1} & \textbf{84.2} & \textbf{47.5} & \textbf{91.0} & \textbf{70.3} & \textbf{81.3} & \textbf{40.6} & \cellcolor{blue!5!white}{\textbf{64.7}} \\
\midrule
\multirow{2}{*}{\texttt{e5-small}} & Base & 23.4 & 77.0 & 38.5 & 86.3 & 60.8 & 70.4 & 29.1 & \cellcolor{blue!5!white}{55.1} \\
 & + FE & \textbf{30.4} & \textbf{77.2} & \textbf{39.0} & \textbf{88.4} & \textbf{64.8} & \textbf{75.3} & \textbf{34.8} & \cellcolor{blue!5!white}{\textbf{58.6}} \\
\midrule
\multirow{2}{*}{\texttt{gte-base}} & Base & 37.0 & \textbf{81.0} & 28.8 & \textbf{92.2} & \textbf{73.8} & 77.1 & \textbf{41.7} & \cellcolor{blue!5!white}{61.7} \\
 & + FE & \textbf{39.3} & 80.7 & \textbf{29.8} & 92.0 & 73.5 & \textbf{79.9} & 40.4 & \cellcolor{blue!5!white}{\textbf{62.2}} \\
\midrule
\multirow{2}{*}{\texttt{gte-large}} & Base & 41.2 & \textbf{83.0} & 31.2 & 92.0 & \textbf{73.6} & 79.3 & \textbf{41.8} & \cellcolor{blue!5!white}{63.2} \\
 & + FE & \textbf{42.6} & 82.7 & \textbf{31.7} & \textbf{92.5} & 73.3 & \textbf{81.2} & 40.6 & \cellcolor{blue!5!white}{\textbf{63.5}} \\
\midrule
\multirow{2}{*}{\texttt{snowflake}} & Base & 34.8 & 48.1 & 36.2 & 22.5 & 64.8 & 24.1 & 37.2 & \cellcolor{blue!5!white}{38.3} \\
 & + FE & \textbf{39.6} & \textbf{68.2} & \textbf{36.4} & \textbf{66.2} & \textbf{68.0} & \textbf{43.2} & \textbf{38.5} & \cellcolor{blue!5!white}{\textbf{51.4}} \\
\bottomrule
\end{tabular}
\caption{NDCG@10 (\%) results with GPT-OSS-20B supervision. Base: Base Retriever; + FE: + Feature Elicitation. NEWS, Signal, and Robust denote TREC-NEWS, Signal-1M, and Robust04. Bold denotes the higher score within each retriever.}
\label{tab:lr_result_gpt20b}
\end{table*}

\begin{table*}[p]
\centering
\small
\setlength{\tabcolsep}{4pt}
\begin{tabular}{llcccccccc}
\toprule
\textbf{Retriever} & \textbf{Condition} & \textbf{BRIGHT} & \textbf{NQ} & \textbf{MuTual} & \textbf{NEWS} & \textbf{Signal} & \textbf{Robust} & \textbf{ArguAna} & \cellcolor{blue!5!white}{\textbf{Avg.}} \\
\midrule
\multirow{2}{*}{\texttt{e5-base}} & Base & 30.8 & 81.9 & 40.3 & 90.3 & 68.1 & 76.5 & 32.9 & \cellcolor{blue!5!white}{60.1} \\
 & + FE & \textbf{36.1} & \textbf{82.3} & \textbf{41.4} & \textbf{90.5} & \textbf{71.4} & \textbf{78.9} & \textbf{39.6} & \cellcolor{blue!5!white}{\textbf{62.9}} \\
\midrule
\multirow{2}{*}{\texttt{e5-large}} & Base & 31.5 & 83.3 & \textbf{47.1} & 90.4 & 66.8 & 77.3 & 34.2 & \cellcolor{blue!5!white}{61.5} \\
 & + FE & \textbf{38.4} & \textbf{84.0} & 46.0 & \textbf{91.5} & \textbf{71.0} & \textbf{81.4} & \textbf{40.2} & \cellcolor{blue!5!white}{\textbf{64.6}} \\
\midrule
\multirow{2}{*}{\texttt{e5-small}} & Base & 23.4 & 77.0 & 38.5 & 86.3 & 60.8 & 70.4 & 29.1 & \cellcolor{blue!5!white}{55.1} \\
 & + FE & \textbf{29.4} & \textbf{77.1} & \textbf{38.8} & \textbf{87.0} & \textbf{64.7} & \textbf{75.1} & \textbf{35.4} & \cellcolor{blue!5!white}{\textbf{58.2}} \\
\midrule
\multirow{2}{*}{\texttt{gte-base}} & Base & 37.0 & \textbf{81.0} & 28.8 & \textbf{92.2} & \textbf{73.8} & 77.1 & \textbf{41.7} & \cellcolor{blue!5!white}{61.7} \\
 & + FE & \textbf{39.0} & 80.8 & \textbf{29.7} & 91.9 & 73.5 & \textbf{80.2} & 40.8 & \cellcolor{blue!5!white}{\textbf{62.3}} \\
\midrule
\multirow{2}{*}{\texttt{gte-large}} & Base & 41.2 & \textbf{83.0} & 31.2 & 92.0 & \textbf{73.6} & 79.3 & \textbf{41.8} & \cellcolor{blue!5!white}{63.2} \\
 & + FE & \textbf{42.5} & 82.8 & \textbf{31.5} & \textbf{92.7} & 73.2 & \textbf{81.7} & 41.3 & \cellcolor{blue!5!white}{\textbf{63.7}} \\
\midrule
\multirow{2}{*}{\texttt{snowflake}} & Base & 34.8 & 48.1 & 36.2 & 22.5 & 64.8 & 24.1 & 37.2 & \cellcolor{blue!5!white}{38.3} \\
 & + FE & \textbf{39.1} & \textbf{69.5} & \textbf{36.4} & \textbf{65.9} & \textbf{68.2} & \textbf{44.1} & \textbf{38.7} & \cellcolor{blue!5!white}{\textbf{51.7}} \\
\bottomrule
\end{tabular}
\caption{NDCG@10 (\%) results with Qwen3-32B supervision. Base: Base Retriever; + FE: + Feature Elicitation. NEWS, Signal, and Robust denote TREC-NEWS, Signal-1M, and Robust04. Bold denotes the higher score within each retriever.}
\label{tab:lr_result_qwen3}
\end{table*}

\twocolumn[{
\section{Additional Examples}
\label{app:examples}
\begin{minipage}{\textwidth}
\centering
\begingroup
\captionsetup{type=table}
\small
\definecolor{CaseElicitation}{RGB}{239,228,244}
\newcommand{\caseactive}{\ensuremath{\bullet}}
\newcommand{\caseinactive}{\ensuremath{\circ}}
\newcommand{\casena}{\textemdash}
\setlength{\tabcolsep}{5pt}
\renewcommand{\arraystretch}{1.2}
\newcommand{\casetext}[2]{%
  \multicolumn{5}{p{\dimexpr\textwidth-2\tabcolsep\relax}}{\textbf{[#1]}~\textit{#2}}\\}
\begin{tabularx}{\textwidth}{p{0.115\textwidth}>{\raggedright\arraybackslash}X c c c}
\toprule
\textbf{Method} & \textbf{Shared feature description} & \textbf{Query} & \multicolumn{2}{c}{\textbf{Document}} \\
\cmidrule(lr){4-5}
 & & & Original & Elicited \\
\midrule
\multicolumn{5}{l}{\textbf{(a) Circle geometry}} \\
\casetext{Query}{Let $ABC$ be a triangle inscribed in circle $\omega$. Let the tangents to $\omega$ at $B$ and $C$ intersect at point $D$, and let $\overline{AD}$ intersect $\omega$ at $P$. If $AB=5$, $BC=9$, and $AC=10$, $AP$ can be written as the form $\frac{m}{n}$, where $m$ and $n$ are relatively prime integers. Find $m + n$.}
\casetext{Document}{Circles $\omega_1$ and $\omega_2$ intersect at points $X$ and $Y$. Line $\ell$ is tangent to $\omega_1$ and $\omega_2$ at $A$ and $B$, respectively, with line $AB$ closer to point $X$ than to $Y$. Circle $\omega$ passes through $A$ and $B$ intersecting $\omega_1$ again at $D \neq A$ and intersecting $\omega_2$ again at $C \neq B$. The three points $C$, $Y$, $D$ are collinear, $XC = 67$, $XY = 47$, and $XD = 37$. Find $AB^2$. $[\ldots]$ By the radical axis theorem $AD, XY, BC$ are concurrent, say at $P$. Moreover, $\triangle DXP \sim \triangle PXC$ by simple angle chasing. $[\ldots]$}
\addlinespace[3pt]
\textbf{SAE} & Step-by-step guides for technical tasks
 & \caseactive & \caseactive & \casena \\
\textbf{Xetrieval} & Circle tangency
 & \caseactive & \caseactive & \caseactive \\
 & Line--circle incidence relations
 & \caseactive & \caseactive & \caseactive \\
\rowcolor{CaseElicitation}
 & Triangles in circle configurations
 & \caseactive & \caseinactive & \caseactive \\
\rowcolor{CaseElicitation}
 & Geometric constraints on segment lengths
 & \caseactive & \caseinactive & \caseactive \\
\midrule
\multicolumn{5}{l}{\textbf{(b) Coordinate geometry}} \\
\casetext{Query}{Let $u$ and $v$ be integers satisfying $0 < v < u$. Let $A = (u,v)$, let $B$ be the reflection of $A$ across the line $y = x$, let $C$ be the reflection of $B$ across the y-axis... The area of pentagon $ABCDE$ is $451$. Find $u + v$.}
\casetext{Document}{$[\ldots]$ Find the area of hexagon $ABCDEF$. Let $R$ be the origin. Noticing that the triangle is a 3-4-5 right triangle, we can see that $A=(4,12), B=(16,3), C=(15,0), D=(5,0), E=(0,5)$, and $F=(0,10)$. Using the shoelace theorem, the area is $\boxed{120}$. $[\ldots]$}
\addlinespace[3pt]
\textbf{SAE} & Systematic problem-solving with step-by-step logic
 & \caseactive & \caseactive & \casena \\
\textbf{Xetrieval} & Coordinate representations of polygons
 & \caseactive & \caseactive & \caseactive \\
\rowcolor{CaseElicitation}
 & Locating vertices from geometric constructions
 & \caseactive & \caseinactive & \caseactive \\
\rowcolor{CaseElicitation}
 & Polygon area from vertex coordinates
 & \caseactive & \caseinactive & \caseactive \\
\midrule
\multicolumn{5}{l}{\textbf{(c) Hardware interfaces}} \\
\casetext{Query}{Custom hardware interface type: I would like to write a controller that needs all joint states to update a single joint. My idea was to create a class MyStateInterface which inherits from hardware\_interface::StateInterface... I want to know if there is a way to pass a class full of control data...}
\casetext{Document}{There was a problem hiding this comment... $[\ldots]$ Suggested change: virtual std::vector<InterfaceDescription> export\_state\_interface\_description()... $[\ldots]$}
\addlinespace[3pt]
\textbf{SAE} & Step-by-step tech how-to guides with specific instructions and links
 & \caseactive & \caseactive & \casena \\
 & Code reviews with technical feedback and suggestions
 & \caseactive & \caseactive & \casena \\
\textbf{Xetrieval} & State-interface APIs in C++
 & \caseactive & \caseactive & \caseactive \\
 & Structured data in interface APIs
 & \caseactive & \caseactive & \caseactive \\
\rowcolor{CaseElicitation}
 & Interface extension and customization
 & \caseactive & \caseinactive & \caseactive \\
\bottomrule
\end{tabularx}
\vspace{4pt}
{\footnotesize
\caseactive\ active\qquad \caseinactive\ inactive\qquad \casena\ not applicable.\par}
\caption{Shared semantic features for additional query--document pairs, with document excerpts. \emph{Elicited} denotes activation in at least one auxiliary document view; shading marks query overlap identified only through these views.}
\label{tab:cases_appendix}
\endgroup
\end{minipage}

\vspace{\textfloatsep}
}]

\noindent Table~\ref{tab:cases_appendix} contrasts SAE's descriptions of procedural or expository form with Xetrieval's descriptions of the semantic relations shared by each query--document pair.
In the circle-geometry example, these relations concern tangency, line--circle incidence, and constraints on segment lengths within triangle and circle configurations.
The coordinate-geometry example links vertex construction and coordinate representations to polygon area.
In the hardware-interface example, the features connect the query's request for a custom state interface to the document's proposed C++ interface changes through structured data and interface customization.

These examples illustrate how Xetrieval moves beyond broad characterizations of textual form to explain the specific semantic connections between queries and documents. The elicited views make relations already supported by document content explicit as shared features, providing a more informative account of document relevance.

\end{document}